\documentclass[letterpaper, 10 pt,conference]{ieeeconf}

\IEEEoverridecommandlockouts  
\usepackage{cite}
\usepackage[pdftex]{graphicx}
\usepackage{lipsum}
\usepackage{float}
\graphicspath{{./Numericals/}}
\DeclareGraphicsExtensions{.pdf}
\usepackage{amsmath}
\usepackage{xurl}
\usepackage{makecell}

\usepackage[cmintegrals]{newtxmath}
\usepackage{color}

\usepackage{algpseudocode,algorithm,algorithmicx}

\usepackage{array}
\usepackage{multirow}

\usepackage[labelformat=simple]{subfig}

\usepackage{hyperref}

\usepackage{tabularx}

\begin{document}

\title{
Can LLMs Understand Social Norms in Autonomous Driving Games?
}

\author{Boxuan Wang, Haonan Duan, Yanhao Feng, Xu Chen, Xuan~Di, Yongjie Fu, Zhaobin Mo~\IEEEmembership{Member,~IEEE} 
\thanks{Boxuan Wang is with the
Department of Mechanical Engineering, Columbia University, New York City, NY 10027 USA (e-mail: bw2812@columbia.edu).}
\thanks{Haonan Duan is with the Data Science Institute, Columbia University, New York City, NY 10027 USA (e-mail: hd2545@columbia.edu).}
\thanks{Yanhao Feng is with the Department of Statistics, Columbia University, New York City, NY 10027 USA (e-mail: yf2684@columbia.edu).}
\thanks{Xu Chen is with the
Department of Civil Engineering and Engineering Mechanics, Columbia University, New York City, NY 10027 USA (e-mail: xc2412@columbia.edu).}
\thanks{Xuan Di is with the Department of Civil Engineering and Engineering Mechanics, Columbia University, New York, NY, 10027 USA, and also with the Data Science Institute, Columbia University, New York, NY, 10027 USA (e-mail: sharon.di@columbia.edu).}
\thanks{Yongjie Fu is with the
Department of Civil Engineering and Engineering Mechanics, Columbia University, New York City, NY 10027 USA (e-mail: yf2578@columbia.edu).}
\thanks{Zhaobin Mo is with the
Department of Civil Engineering and Engineering Mechanics, Columbia University, New York City, NY 10027 USA (e-mail: zm2302@columbia.edu).}
}

\maketitle

\begin{abstract}
Social norm is defined as a shared standard of acceptable behavior in a society. The emergence of social norms fosters coordination among agents without any hard-coded rules, which is crucial for the large-scale deployment of  autonomous vehicles (AVs) in an intelligent transportation system. This paper explores the application of large language models (LLMs) in understanding and modeling social norms in autonomous driving games. We introduce LLMs into autonomous driving games as intelligent agents who make decisions according to text prompts. These agents are referred to as LLM agents. Our framework involves LLM agents playing Markov games in a multi-agent system (MAS), allowing us to investigate the emergence of social norms among individual agents. We aim to identify social norms by designing prompts and utilizing LLMs on textual information related to the environment setup and the observations of LLM agents. Using the OpenAI Chat API powered by GPT-4.0, we conduct experiments to simulate interactions and evaluate the performance of LLM agents in two driving scenarios: unsignalized intersection and highway platoon. The results show that LLM agents can handle dynamically changing environments in Markov games, and social norms evolve among LLM agents in both scenarios. In the intersection game, LLM agents tend to adopt a conservative driving policy when facing a potential car crash. The advantage of LLM agents in games lies in their strong operability and analyzability, which facilitate experimental design.
\medskip

\end{abstract}








\IEEEpeerreviewmaketitle

\section{Introduction}
\label{sec:intro}

Large Language Models (LLMs) have demonstrated remarkable capabilities in natural language understanding, enabling them to process and generate human-like text across a wide range of domains \cite{ruan2024speech,yin2024fine,fu2024vlm,bai2024beyond,bai2024gradient}. The increasing popularity of LLMs facilitates the design of numerous applications, allowing LLMs to frequently interact with us in our daily lives. With enhanced capabilities, LLMs hold considerable promise for applications in traffic planning and autonomous driving. This paper aims to study how LLM-guided agents behave and interact with each other in autonomous driving scenarios. Specifically, we utilize LLMs to analyze autonomous driving games in which LLM-guided agents are tasked with making decisions in simulated driving environments. Our goal is to investigate whether these agents can understand social norms in autonomous driving games.

In recent years, autonomous driving technology has witnessed significant advancements in recent years, promising safer and more efficient transportation systems. However, the integration of autonomous vehicles into society \cite{chen2023ssd} raises concerns about their interaction with human drivers, pedestrians, and other road users. One critical issue of this interaction is the adherence to social norms, the informal rules that guide vehicles to navigate the road. Understanding and effectively modeling these social norms are essential for the development of autonomous driving systems capable of navigating real-world scenarios effectively.

In this paper, we explore the potential of leveraging Large Language Models (LLMs) to understand social norms in autonomous driving games. By utilizing LLMs on textual information related to autonomous driving scenarios, we can teach them to identify the social norms among vehicles on the road. By observing and analyzing the choices made by players in these games, we can extract implicit social norms that regulate their behavior. Our research aims to contribute to the development of socially aware autonomous driving systems that can effectively navigate complex driving environments. Leveraging LLMs to understand and model social norms can empower safer interactions between autonomous vehicles and other road users.



\subsection{Related work}

To explore LLMs' capability of understanding human behaviors \cite{kwon2023reward,Bai2022TrainingAH,ruan2024llm}, there has been a growing trend of analyzing LLMs in the context of game theory \cite{mao2023alympics,chahine2024large}, including fairness \cite{horton2023llm}, dilemma \cite{akata2023playing} and rationality \cite{brand2023market,yiting2023ration,dillion2023llm}. Many studies employ LLMs to replace humans as research subjects \cite{aher2023llm,argyle2023lan}. To study complex interactions among players in game theory problems, LLMs are introduced into game-theoretic frameworks as intelligent agents, referred to as LLM-based agents \cite{sumers2024cognitive,xu2023exploring}. These LLM-based agents empower the development of sophisticated systems \cite{Ouyang2022TrainingLM,fan2023large,yin2023deep,Peng2024causal} where players' behaviors can be simulated via LLMs. For example, LLM-based agents are utilized to study cooperation and coordination behaviors in the prisoner's dilemma \cite{akata2023playing}. This paper further explores the capability of LLM-based agents regarding social outcomes in games. More specifically, we investigate whether LLM-based agents can form desired social norms in autonomous driving games to improve driving efficiency and road safety. 

Social norms have been widely studied in games. Most literature on social norms primarily focuses on stateless matrix games \cite{delgado2002conventions,sen2007social,villatoro2011norm,yu2013collective,frank2013convention}. To capture dynamic environments,  Markov games are developed to study the emergence of social norms \cite{lerer2019conventions,koster2020convention} in sequential decision making. A learning-based framework is proposed to study how social norms evolve in autonomous driving games \cite{chen2022sl}, where agents are guided by a deep reinforcement learning (DRL) algorithm. Compared to DRL-based agents, LLM-based agents have the following advantages: strong operability, relatively simple experimental design in game theory, and strong analyzability \cite{fan2023large}. Therefore, we introduce LLM-based agents into autonomous driving games to handle dynamically changing driving environments.

\subsection{Contributions of this paper}
This work focuses on how LLMs can facilitate navigation in autonomous driving games. Our study utilizes the OpenAI Chat API \cite{openai2024gpt4}, powered by GPT-4.0, to conduct experiments. Our contributions include: 
\begin{enumerate}
    \item We propose a framework where LLM agents play Markov games in a multi-agent system (MAS) in order to investigate whether social norms emerge among individual agents. 
    \item We design prompts to simulate interactions in autonomous driving games. The prompts consist of ``system" messages outlining the general settings (e.g., the road layout) and ``user" messages regarding the observations and decision making of LLM agents.
    \item We apply the proposed framework to two traffic scenarios: the unsignalized intersection and highway platoon, to examine social norms formed by LLM agents. 
\end{enumerate}



The rest of the paper is organized as follows: In Section \ref{sec:prelim}, we first present preliminaries regarding Markov game, LLMs and social norm. Section \ref{sec:method} introduces our framework in which LLM agents play Markov games via prompt design. Section \ref{sec:experiments} presents numerical experiments and results. Section \ref{sec:conclu} concludes.

\begin{figure*}
    \centering
        \vspace{-.1in}
    \includegraphics[width=0.8\textwidth]{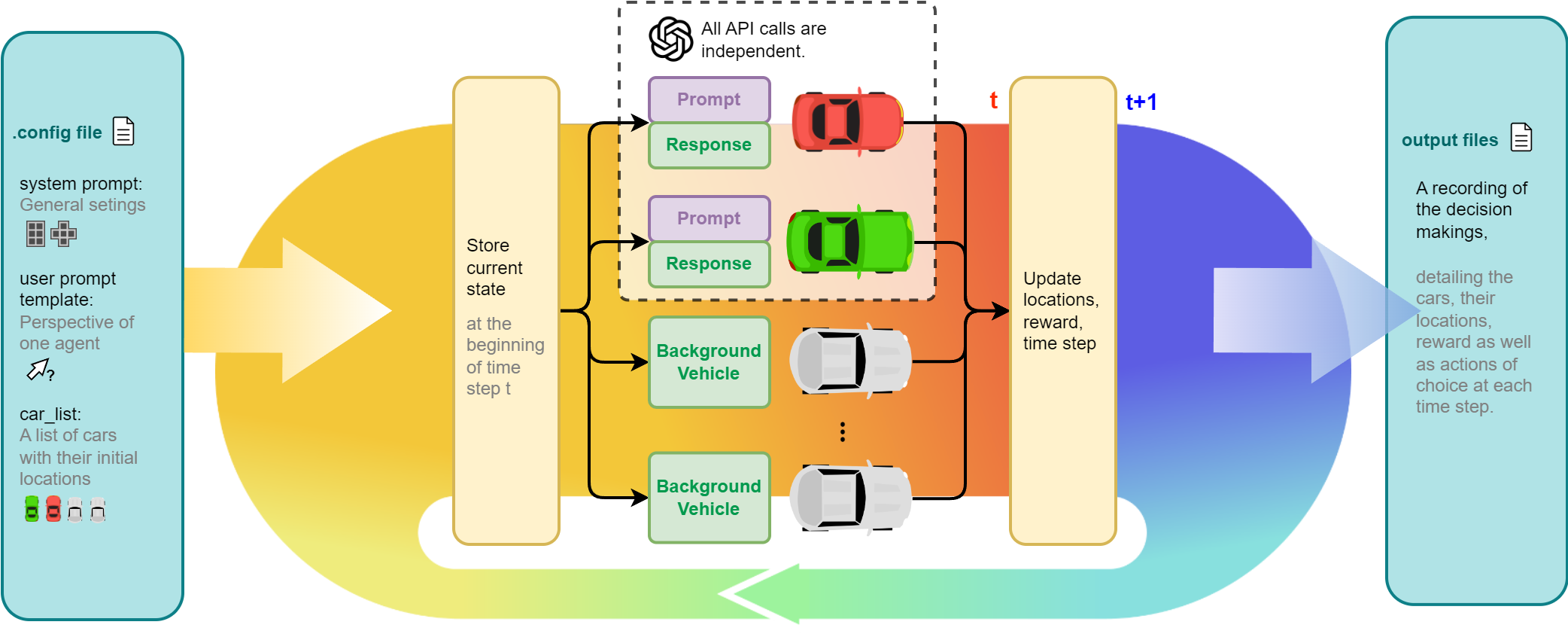}
        \vspace{-.05in}
    \caption{LLM for autonomous driving games}
    \label{fig:flow}
        \vspace{-.2in}
\end{figure*}

\section{Preliminary}
\label{sec:prelim}
\subsection{Markov Game} 
\label{sec:markov_game}

\begin{figure}[H]
    \centering
    \vspace{-.3in}
    \subfloat[\scriptsize{Scenario 1}]{\includegraphics[width=0.39\linewidth]{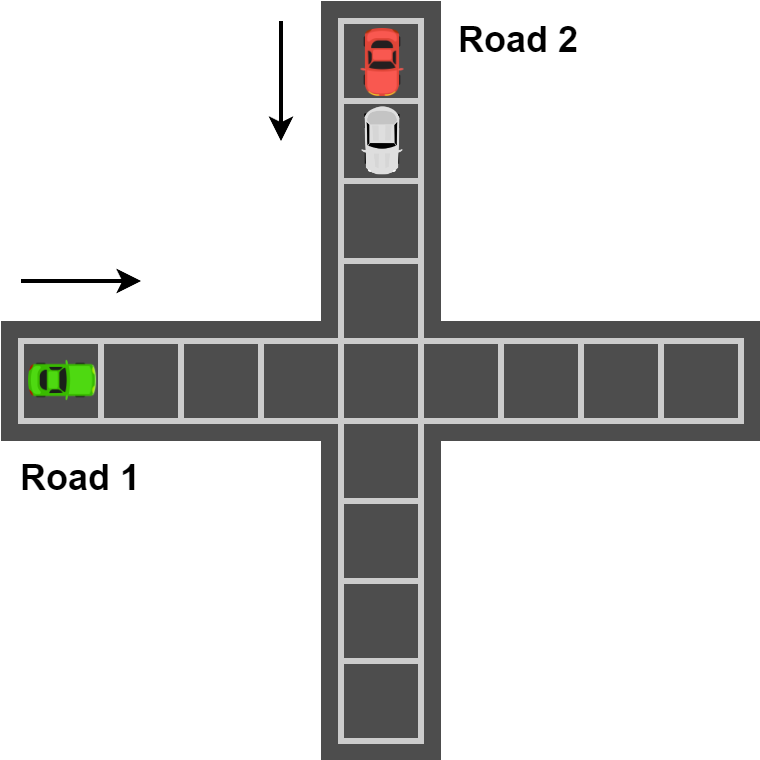}\label{fig:setup_int}}
    \hspace{.3in}
    \subfloat[\scriptsize{Scenario 2}]{\includegraphics[width=0.18\linewidth]{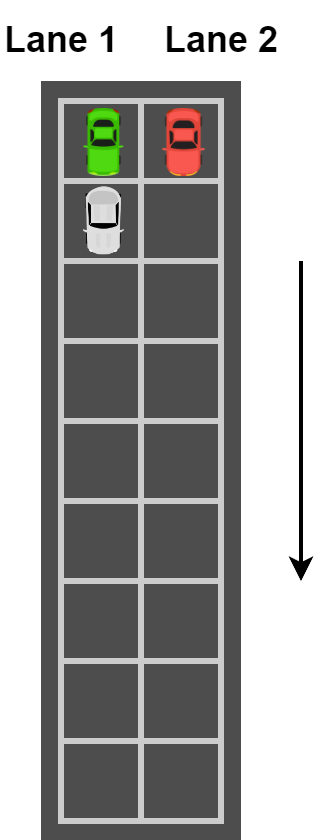}\label{fig:setup_plat}} 
    \vspace{-.1in}
    \caption{Game scenarios}
    \label{fig:setup_}
    \vspace{-.1in}
\end{figure}

A Markov game, also known as a stochastic game, is a mathematical framework that generalizes both Markov Decision Processes (MDPs) and game theory to model dynamic interactions among rational players in uncertain environments. In its general form, it is defined by a state set $\mathcal{S}$ and action sets $\mathcal{A}_1,...,\mathcal{A}_n$ for each corresponding agent in the environment \cite{michael1994markov}. Each agent also has a corresponding reward function $\mathcal{R}_i: \mathcal{S} \times \mathcal{A}_i \rightarrow \mathbb{R}$ for agent $i$. The goal of the Markov Game is to find an optimal policy for each agent. In this paper, we investigate the behavior and decision-making of LLMs when they play the Markov Game as agents. We use $\mathcal{S}$ to denote the overall environment, $\mathcal{A}_i$ to denote the action set of agent $i$, $\mathcal{O}_i$ to denote the observation set of agent $i$, $\mathcal{R}_i$ to denote the current reward of agent $i$. LLM chooses one of the actions from $\mathcal{A}_i$, with $\mathcal{O}_i$ and partial $\mathcal{S}$ as input. In this paper, the action choosing policy of an agent was solely decided by LLMs. Our primary focus is on two games as follows:

\emph{\textbf{Scenario 1-Unsignalized intersection}}: The environment $\mathcal{S}$ is visualized in Figure \ref{fig:setup_int}, with Road 1 extending from the West to the East and Road 2 extending from the North to the South. The green and red cars represent the agents traveling on Road 1 and Road 2, respectively. Agents aim to cross the intersection, reach the end of the roads, and maximize their cumulative rewards. Table \ref{tab:setup} summarizes the setup of agent $i$ at the beginning of time $t$. The location of agent $i$, $x_{i,t}$ corresponds to the cell it currently occupies. At each time step, each agent can choose "Go" or "Stop", each with different action reward. The observation $o_{i, t}$ is a subset of current state space $\mathcal{S}$. Background vehicles (marked in white) with predetermined driving strategies are non-strategic players in the driving environment. The game terminates when all agents complete their trips or a crash occurs.   

\emph{\textbf{Scenario 2-Highway platoon}}: The environment $\mathcal{S}$ consists of two lanes, shows in Figure \ref{fig:setup_plat}, going from North to South. Red and green agents each occupy one lane. Agents aim to reach the end of the lanes and maximize their cumulative rewards. In addition to the "Go" and "Stop" action, agents can now also choose to switch lanes. We defined the success of the game as forming a platoon at least once and completing the trips without any car crashes. A platoon is characterized by the presence of cars in the same lane. With the incentive of the reward and the overall goal, the agents will aim to complete their trips as quickly as possible, while avoiding accidents and attempting to maintain platoons. Table \ref{tab:setup} summarize the setup of the  platoon game.

\begin{table}[H]
\centering
\small
\vspace{-.2in}
    \begin{tabular}{|c|c|}
        \hline
        \multicolumn{2}{|c|}{Scenario 1}
        \\ \hline
        $s_{i,t}\in \mathcal{S}$ & Location $x_{i,t}$ and cumulative reward $\rho_{i,t}$ \\ \hline
        $o_{i,t}\in \mathcal{O}$ & Observation of the current environment \\ \hline
        $a_{i,t}\in \mathcal{A}$ & Go: move forward by one cell; Stop: no move \\ \hline  
        $r_{i,t}\in \mathcal{R}$ & Go: -2; Stop: -2; Crash: -5 \\ \hline
        \multicolumn{2}{|c|}{Scenario 2}
        \\ \hline
        $s_{i, t}\in \mathcal{S}$ & Location $x_{i,t}$ and cumulative reward $\rho_{i,t}$ \\ \hline
        $o_{i, t}\in \mathcal{O}$ & Observation of the current state space \\ \hline
        $a_{i, t}\in \mathcal{A}$ & Go; Stop; Lane change: switch to the other lane \\ \hline  
        $r_{i, t}\in \mathcal{R}$ & Go; Stop; Lane: -4; Crash: -5; Platoon: +2 \\ \hline
    \end{tabular}
    \vspace{-.05in}
    \caption{The setup of game scenarios}
    \label{tab:setup}
    \vspace{-.1in}
\end{table}

\subsection{Large Language Models} 
\label{sec:llm}

A large language model (LLM) is a type of artificial intelligence algorithm designed to understand, generate, and interact with human language at a large scale \cite{openai2024gpt4}. These models are trained on vast amounts of text data, learning patterns, vocabulary, grammar, and even nuances of language such as context, tone, and implications. They use a specific architecture known as Transformer \cite{vaswani2023attention}, which allows them to efficiently process sequences of words and predict the next word in a sentence, understand the meaning of a text, or generate new text that follows a given prompt \cite{hadi2023large}. LLMs are highly versatile AI tools with applications spanning multiple industries and fields \cite{mohapatra2023exploring}. They excel in tasks such as content creation, customer support, education, language translation, software development, legal documentation, healthcare research, optimization, sentiment analysis, and accessibility technologies. In this paper, we examine the application of LLMs in autonomous driving games to see if how they make decisions and form social norms.

\subsection{Social Norms} 
\label{sec:social_norms}

Social norms represent the collective expectations regarding appropriate conduct within groups \cite{nyborg2016social}. They encompass both the unwritten agreements that dictate how society members should act, and the formalized regulations and laws that are established. Social norms in car driving \cite{chen2022sl} are the unwritten rules that guide driver behavior on the road. Our primary focus is on examining whether expected social norms can evolve among LLM agents. The expected social norm in Scenario 1 is ``yielding to others" at the intersection. The expected social norm in Scenario 2 is the formation of a platoon among agents.




\section{Methodology}
\label{sec:method}

In this section, we present the framework of leveraging LLMs to understand social norms in Markov games. 

In Figure \ref{fig:flow}, we utilize prompt-chaining for LLM interactions in the Markov game. At each time step, the program invoked the chat model from the perspective of each individual agent. The games are submitted to LLMs as prompts in
which the respective game, including the choice options, is described. Once LLMs have made their choices (i.e., the actions made by strategic and non-strategic agents), which we track as a completion of the given text, we update the prompts and then submit the new prompt to LLMs for the next state in the game. The textual information includes the reward and each agent's current location. When a car completes its trip, we remove it from the game, meaning that it is no longer observed by other cars, or prompted about its actions. Its location and reward stop updating as well. This prevents any potential interference with the cars still in the game. The output includes the decision making, rewards and states of agents in the Markov game, which is used to analyze social norms.

\begin{figure}[h]
    \centering
        \vspace{-.1in}
    \includegraphics[width=0.6\linewidth]{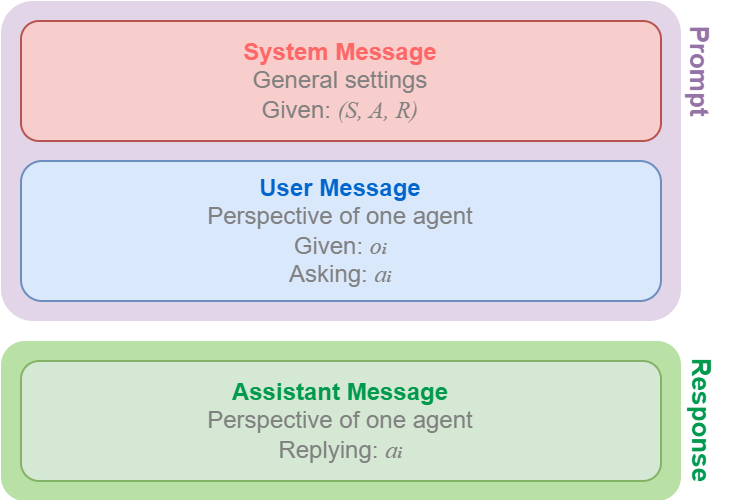}
        \vspace{-.05in}
    \caption{Prompt design}
    \label{fig:enter-label}
        \vspace{-.1in}
\end{figure}

\emph{\textbf{Prompt design}}: The chat model facilitates conversations through three distinct roles, including ``user", ``assistant", and ``system" (See Figure \ref{fig:enter-label}). ``User" represents the human participant interacting with the AI. The ``assistant" role embodies the AI providing responses to the user’s prompts. Finally, the ``system" role is primarily responsible for defining the assistant's behavior—this can involve setting a specific language, character, or viewpoint for the assistant. The chat model prompt is structured as a series of messages, each identified by the role of the sender and the message content. By concatenating past interactions from both the user and assistant, the model can more effectively engage in multi-turn conversations. The detailed prompts are in Section \ref{sec:experiments}. In our research, we assume that the agents will not be capable of making predictions based on any historical context regarding previous states. This leads us to use only one ``system" message and one ``user" message to construct our prompt. The ``system" message outlines the general settings, defining parameters like the environment space $\mathcal{S}$ (i.e., the road layout), action space $\mathcal{A}$ (i.e., all the permitted moves), and the reward function $\mathcal{R}$. Meanwhile, the ``user" message takes the perspective of a specific car. This message first identifies agent $i$ (i.e., its color) and its present observation $o_i$ (i.e., the locations of all the cars and its own cumulative reward). After that, the prompt asks the agent about the action $a_i$ it should take and the resulting location due to the action of choice.

We employ the OpenAI Chat API, specifically the GPT-4.0 model, to conduct our experiments. Notably, GPT-4.0 has demonstrated significant improvement over its predecessor, GPT-3.5, by at least 20\%, as per OpenAI's assertions, on an array of benchmarks such as the SuperGLUE language comprehension and the LAMBADA linguistic modeling tests. As an advanced version of the transformer-based language model \cite{openai2024gpt4}, GPT-4.0 holds considerable potential for traffic planning and autonomous driving. 


Note that our focus is not cooperative game. Therefore, we ensure that when an agent made a decision, it does not have access to the concurrent decisions made by others. We do not ask the chat model to simulate the entire game in one prompted response, as it has access to the private observations of all agents and can implicitly use other agents' private observations to make decisions. Instead, we simulate the traffic in a turn-based manner, where each turn represents a time step. Each inquiry to the chat model could only take the perspective of one agent. To further prevent access to concurrent decisions made by other agents, the game state is updated at the end of each time step, after receiving responses from all agents.

\section{Experiments}
\label{sec:experiments}

In this section, we conduct experiments on two game scenarios to understand how social norms can be formed when LLMs play Markov games.


\textbf{Experiment setup}: The environments of both the intersection and platoon scenarios adopt a grid representation consisting of multiple cells, with each cell assigned a pair of coordinates to denote the location of vehicles. The red and green cars are strategic agents, while the white car serves as a background vehicle. All background vehicles are not involved in the formation of social norms. The setup is stored in a configuration containing all prompts, a list of cars, and their initial positions. Given that the general settings is consistent for all agents and is unaffected by time, the system message portion of the prompt is stored as a static string. On the other hand, each car’s private observation, influenced by time and specific agent conditions, is stored as templates. Each time these templates are accessed, they are customized in accordance with present conditions. Each car could observe both the location and the cumulative reward of itself, but could only see the locations of all other cars. To easily concatenate the observation about all other cars, we store the user message as two separate templates: one for the observing itself, and one for all other cars. 

\begin{figure}[h]
    \centering
    \vspace{-.1in}
    \includegraphics[width=0.7\linewidth]{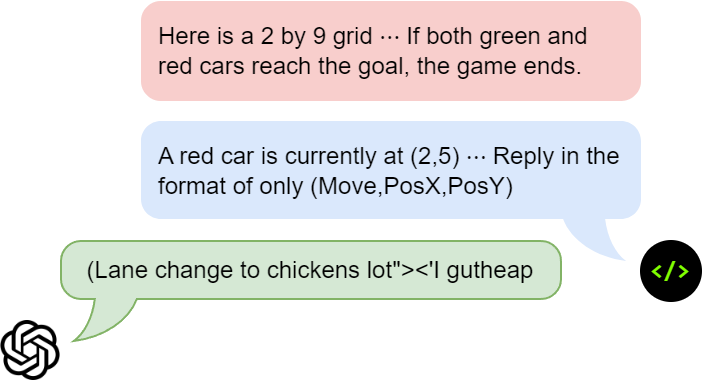}
    \vspace{-.05in}
    \caption{Responses (temperature = 2.0)}
    \label{fig:ex-failed-tmp-2}
    \vspace{-.1in}
\end{figure}

In both scenarios, we utilize the OpenAI API with the specified parameters: model = gpt-4, temperature = 0.1, max token = 10. Increasing the temperature enhances the randomness of the response \cite{openaiapi}.  Our testing reveal that raising the temperature beyond 1.0 sometimes results in unstable behaviors. Further increasing the temperature to 2.0 leads to responses that are essentially unintelligible, as exemplified in Figure \ref{fig:ex-failed-tmp-2}. The max token parameter restricts the maximum length of the response. As per OpenAI's tokenizer, the language model can consistently provide responses in the preferred format within 10 tokens \cite{openaitokenizer}. 

\subsection{Scenario 1-Unsignalized intersection}
\begin{figure}[h]
    \centering
        \vspace{-.1in}
    \includegraphics[width=1\linewidth]{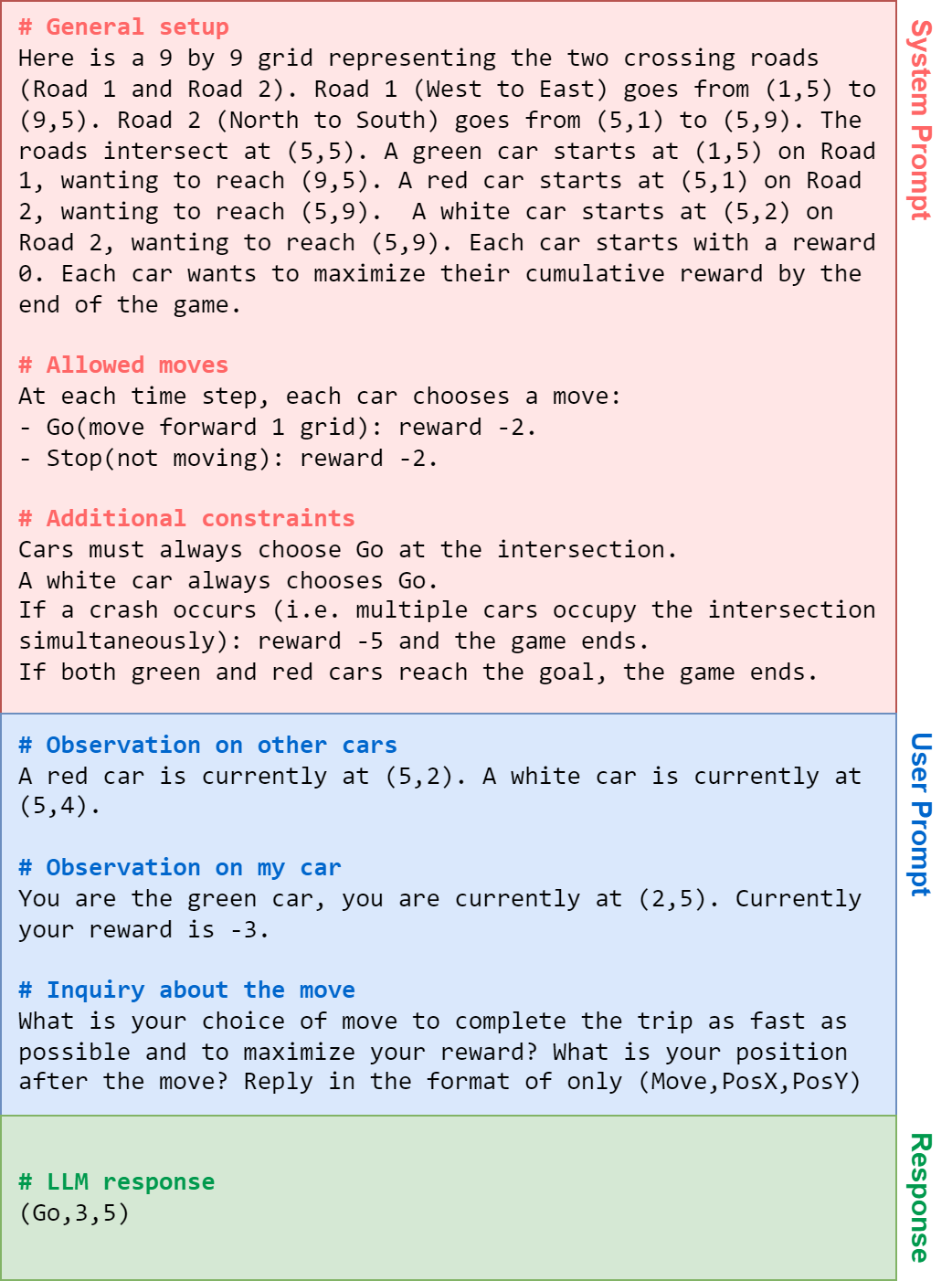}
        \vspace{-.2in}
    \caption{Prompts and responses - Scenario 1}
    \label{fig:ex-inter}
        \vspace{-.1in}
\end{figure}
\begin{table}[h]
\small
\vspace{-.1in}
    \begin{tabular}{|p{0.6in}|p{0.6in}|p{0.6in}|p{0.6in}|}
        \hline
        \# of BVs & \# of tests  & \# of Yields &  Rate \\ \hline
        0 & 50 & 48 & 0.96 \\ \hline
        
        1 & 50 & 50 & 1\\ \hline
        
        2 & 50 & 50 & 1 \\ \hline
        
        3 & 50 & 50 & 1 \\ \hline
        
        4 & 50 & 50 & 1 \\ \hline
    \end{tabular}
    \vspace{-.05in}
\caption{Performance - unsignalized intersection}
\label{tab:bv_inter}
\vspace{-.1in}
\end{table}

Figure \ref{fig:ex-inter} shows an example of our prompts for scenario where the environment of an unsignalized intersection has a 9 by 9 grid. In order to inspect the behavior of LLMs, we simulate the Markov Game 50 times. Table \ref{tab:bv_inter} summarizes the performance of LLMs when varying numbers of background vehicles. The results demonstrate the formation of social norms at the unsignalized intersection (i.e., yielding to others). It is noteworthy that crashes may occur when no background vehicles are present. This can be attributed to the fact that the presence of background vehicles restricts the action space for ego cars, thereby reducing the likelihood of vehicles encounters at the intersection and facing a crash.

\begin{figure}[h]
    \vspace{-.1in}
    \centering
    \includegraphics[width=0.8\linewidth]{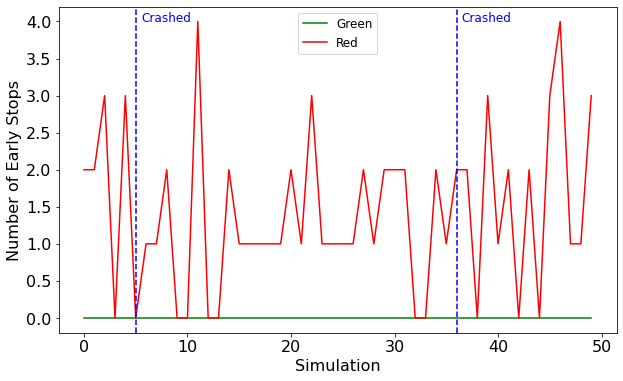}
        \vspace{-.1in}
    \caption{Early stop behaviors}
    \label{fig:early_stop}
        \vspace{-.1in}
\end{figure}

We now look into the case when there are no background vehicles. The results show that the red car tends to early stop before several grids of the intersection. The early stop behavior measures the waiting time of vehicles. It is defined as the stop action of vehicles when they haven't arrived at the intersection (5,5), i.e., vehicles stop at positions other than (4,5) or (5,4), 1 grid away from the intersection.  Figure \ref{fig:early_stop} visualizes the line plot of the number of early stops in each simulation. The x-axis represents the simulation. The y-axis represents the number of early stops in each simulation. The red and green lines denote the early stop behaviors of red and green vehicles, respectively. The crash cases are marked by blue dotted lines. The average number of early stops of the red car is 1.42, and the average of early stops of green car is 0.0. The results show that a conservative driving policy is adopted by the red car when playing the Markov game at the unsignalized intersection.

We then investigate the crash cases, which occurr in the 6th and 37th simulations. In the 6th simulation, the red car does not stop earlier and decides to proceed at the same time as the green car, resulting in a crash. In the 37th simulation, the red car stops earlier, while the green car stops 1 grid away from the intersection. When the red car arrives at the intersection, both cars decide to proceed simultaneously, leading to a crash.

To eliminate the early stop behaviors and shorten the waiting time of vehicles, we increase the reward of Go action from -2 to 0. We re-simulate the Markov game for 50 times. The results show that (1) The average number of early stops for the red car decreases from 1.42 to 0.48, suggesting that the reward design can influence driving behavior. (2) The success rate of adherence to social norms decreases from 0.96 to 0.94 due to an increase in the number of car crashes. It indicates a trade-off between waiting time and road safety in the intersection game. The decrease in waiting time would increase the possibility of crashes.

\subsection{Scenario 2-Highway Platoon}

\begin{figure}[h]
    \centering
    \vspace{-.1in}
    \includegraphics[width=1\linewidth]{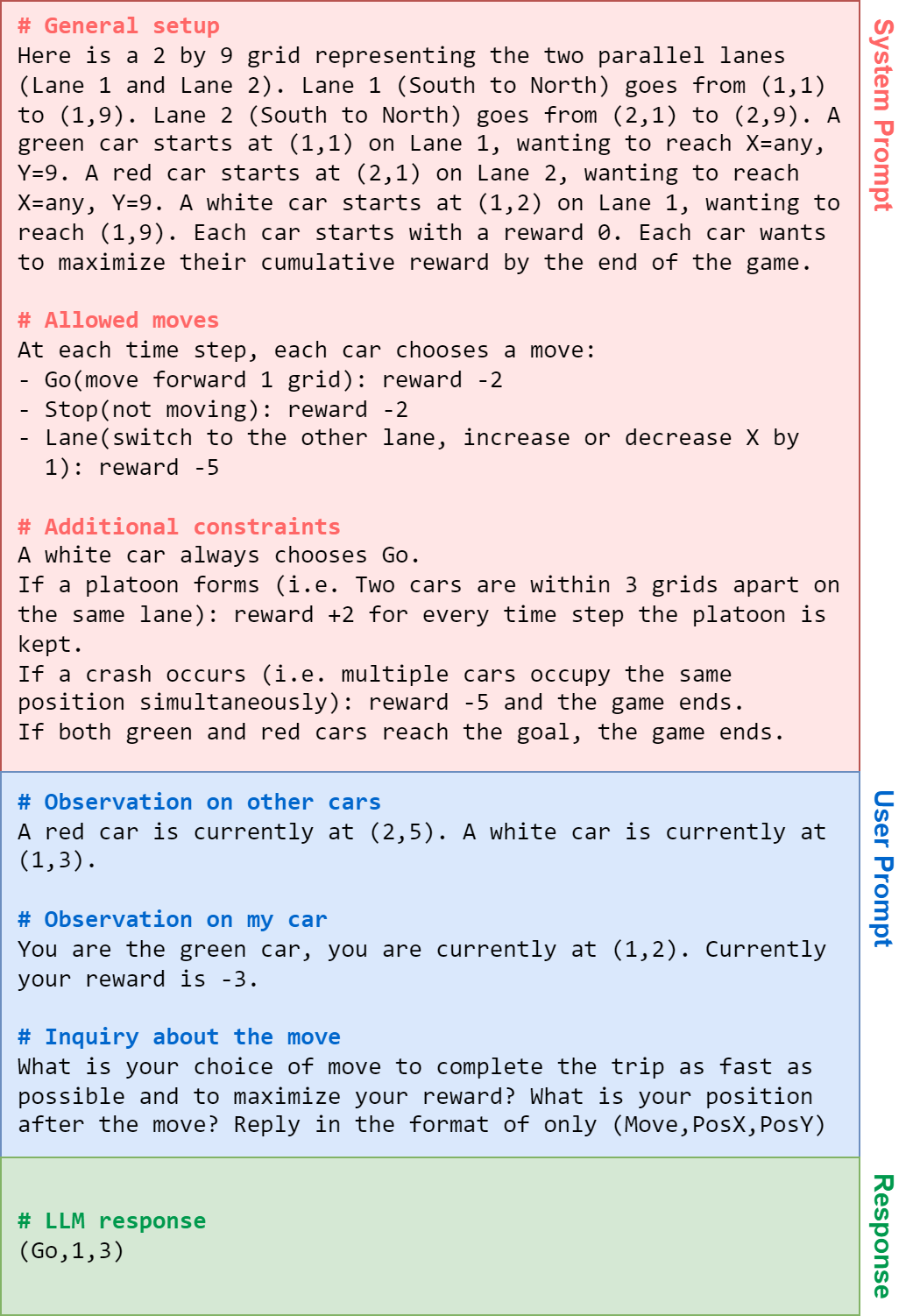}
    \vspace{-.2in}
    \caption{Prompts and responses - Scenario 2}
    \label{fig:ex-plat}
    \vspace{-.05in}
\end{figure}

Figure \ref{fig:ex-plat} shows an example of prompts for scenario where the environment of a highway platoon has a 2 by 9 grid.

\begin{table}[h]
\centering
\small
\vspace{-.05in}
    \begin{tabular}{|p{0.6in}|p{0.6in}|p{0.6in}|p{0.6in}|}
        \hline
        \# of BVs & \# of tests  & \# of Platoons & Rate \\ \hline
        0 & 50 & 48 & 0.96 \\ \hline
        
        1 & 50 & 46 & 0.92 \\ \hline
        
        2 & 50 & 49 & 0.98 \\ \hline
        
        3 & 50 & 43 & 0.86 \\ \hline
        
        4 & 50 & 40 & 0.80 \\ \hline
    \end{tabular}
    \vspace{-.05in}
\caption{Performance - highway}
\label{tab:pla_bv}
\vspace{-.1in}
\end{table}

\begin{figure}[h]
    \centering
    \vspace{-.1in}
    \includegraphics[width=0.7\linewidth]{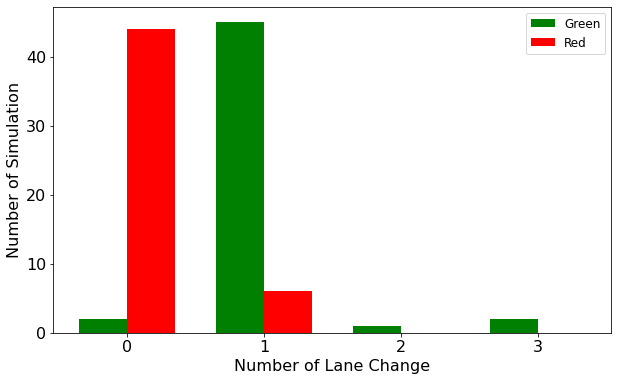}
    \vspace{-.1in}
    \caption{Lane-change frequency}
    \label{fig:lane_change}
    \vspace{-.1in}
\end{figure}

\begin{figure}[h]
    \centering
    \vspace{-.1in}
    \includegraphics[width=0.7\linewidth]{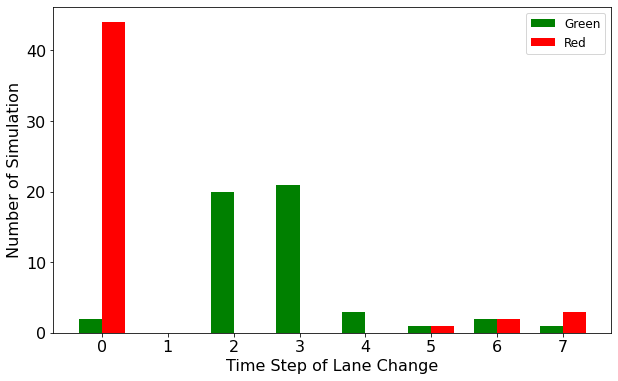}
    \vspace{-.1in}
    \caption{Time step of lane-change}
    \label{fig:lane_time}
        \vspace{-.1in}
\end{figure}

\begin{figure}[h]
    \centering
    \vspace{-.1in}
    \includegraphics[width=0.7\linewidth]{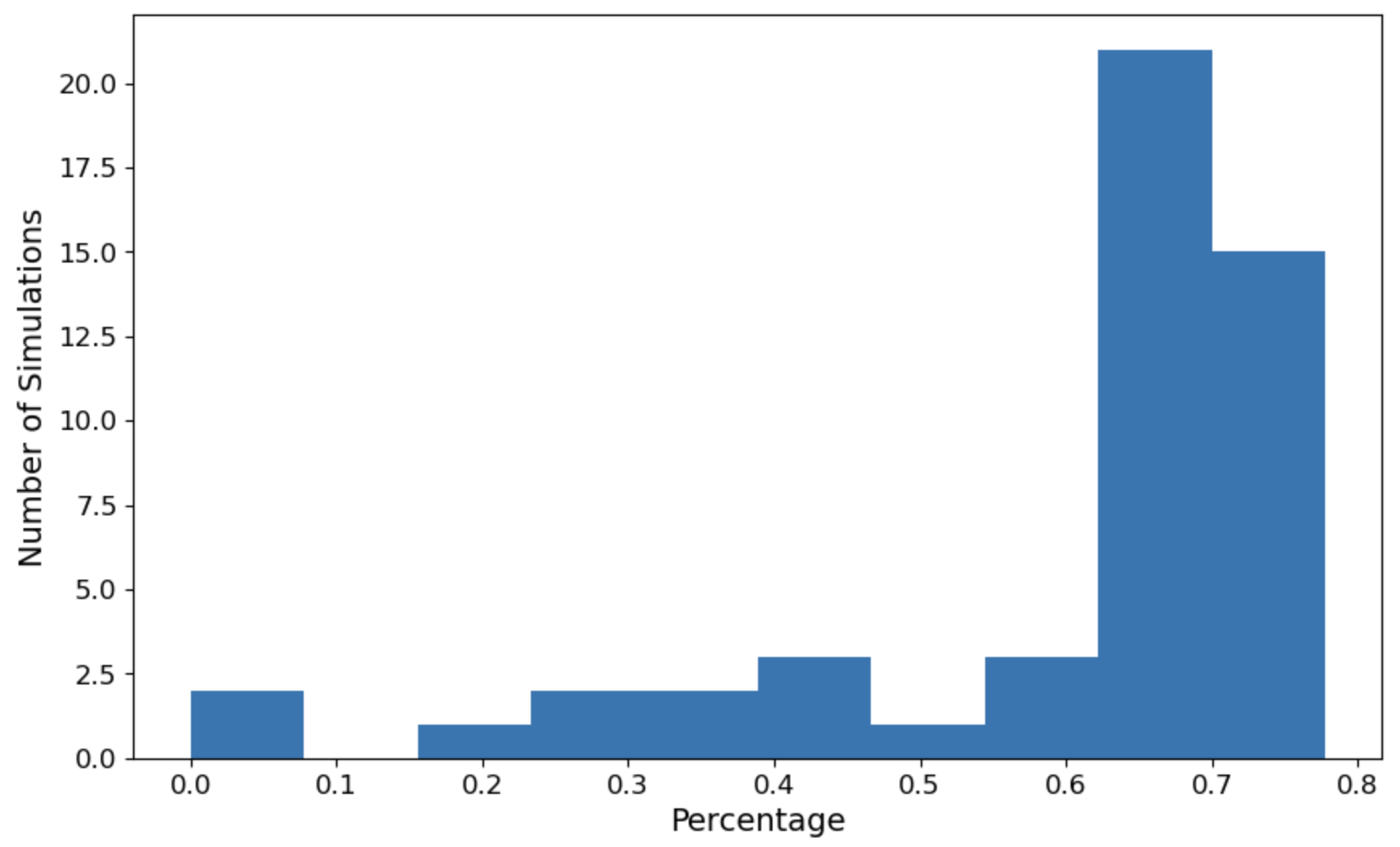}
    \vspace{-.1in}
    \caption{Time length of forming a platoon}
    \label{fig:platoon_time}
    \vspace{-.1in}
\end{figure}

Table \ref{tab:pla_bv} summarizes the performance of LLM agents when varying numbers of background vehicles. The results show that the success rate of forming platoons is high, indicating the emergence of social norms among LLM agents. Figure \ref{fig:lane_change} demonstrates lane-change behaviors. The histogram visualizes the frequency of lane-change actions. The x-axis denotes the number of lane-change actions and the y-axis is the number of simulations. We can tell that there is at most three lane-changes per simulation, and one lane-change in most cases. The green car changes lane more frequently in a comparison to the red car. Figure \ref{fig:lane_time} is the stacked histogram of the time step when the agents choose to change the lane. It is shown that most of the LLM agents change lanes before the fourth time step. It indicates that LLM agents are inclined to change lanes and form platoons at an early stage, demonstrating an awareness that platoon formation is an efficient driving strategy on highways. Figure \ref{fig:platoon_time} plots the histogram of percentage time of platoon in each game. Moreover, in the majority of simulations where platoon is formed, there are more than a 60 percent of the time that two agents forming a platoon on the highway. 

\section{Conclusion And Future Work}
\label{sec:conclu}

This paper has demonstrated the potential of LLMs in understanding and modeling social norms in autonomous driving scenarios. By introducing LLM agents into autonomous driving games, we have observed the emergence of social norms among agents navigating complex driving environments. Our experiments have shown that LLM agents can adapt and conform to social norms, thereby contributing to safer and more efficient driving behaviors. Overall, the advantage of employing LLM agents in autonomous driving games lies in their strong operability and analyzability, which facilitate experimental design and provide valuable insights into the dynamics of social norms in driving environments. By further investigating the behavior of LLM agents in various driving scenarios and refining our experimental framework, we can contribute to the development of more socially aware autonomous driving systems.

This work can be extended in the following ways: (1) We will utilize LLM agents in more complex real-world scenarios, such as sequential social dilemmas. (2) We aim to explore how to construct a unified, controllable, and efficient framework for simulating strategic interactions and facilitating game design using LLMs. (3) We will compare the behaviors of human players and LLM agents to determine the extent to which LLMs can achieve strategic reasoning.

\bibliographystyle{IEEEtran}
\bibliography{cite_boxuan}

\begin{thebibliography}{10}
\providecommand{\url}[1]{#1}
\csname url@samestyle\endcsname
\providecommand{\newblock}{\relax}
\providecommand{\bibinfo}[2]{#2}
\providecommand{\BIBentrySTDinterwordspacing}{\spaceskip=0pt\relax}
\providecommand{\BIBentryALTinterwordstretchfactor}{4}
\providecommand{\BIBentryALTinterwordspacing}{\spaceskip=\fontdimen2\font plus
\BIBentryALTinterwordstretchfactor\fontdimen3\font minus \fontdimen4\font\relax}
\providecommand{\BIBforeignlanguage}[2]{{%
\expandafter\ifx\csname l@#1\endcsname\relax
\typeout{** WARNING: IEEEtran.bst: No hyphenation pattern has been}%
\typeout{** loaded for the language `#1'. Using the pattern for}%
\typeout{** the default language instead.}%
\else
\language=\csname l@#1\endcsname
\fi
#2}}
\providecommand{\BIBdecl}{\relax}
\BIBdecl

\bibitem{ruan2024speech}
K.~Ruan, X.~He, J.~Wang, X.~Zhou, H.~Feng, and A.~Kebarighotbi, ``S2e: Towards an end-to-end entity resolution solution from acoustic signal,'' in \emph{ICASSP 2024 - 2024 IEEE International Conference on Acoustics, Speech and Signal Processing (ICASSP)}, 2024, pp. 10\,441--10\,445.

\bibitem{yin2024fine}
\BIBentryALTinterwordspacing
K.~Yin, C.~Liu, A.~Mostafavi, and X.~Hu, ``Crisissense-llm: Instruction fine-tuned large language model for multi-label social media text classification in disaster informatics,'' 2024. [Online]. Available: \url{https://arxiv.org/abs/2406.15477}
\BIBentrySTDinterwordspacing

\bibitem{fu2024vlm}
Y.~Fu, Y.~Li, and X.~Di, ``Gendds: Generating diverse driving video scenarios with prompt-to-video generative model,'' in \emph{2024 IEEE International Conference on Intelligent Transportation Systems (ITSC)}, 2024.

\bibitem{bai2024beyond}
G.~Bai, Z.~Chai, C.~Ling, S.~Wang, J.~Lu, N.~Zhang, T.~Shi, Z.~Yu, M.~Zhu, Y.~Zhang \emph{et~al.}, ``Beyond efficiency: A systematic survey of resource-efficient large language models,'' \emph{arXiv preprint arXiv:2401.00625}, 2024.

\bibitem{bai2024gradient}
G.~Bai, Y.~Li, C.~Ling, K.~Kim, and L.~Zhao, ``Gradient-free adaptive global pruning for pre-trained language models,'' \emph{arXiv preprint arXiv:2402.17946}, 2024.

\bibitem{chen2023ssd}
X.~Chen, X.~Di, and Z.~Li, ``Social learning for sequential driving dilemmas,'' \emph{Games}, vol.~14, no.~3, 2023.

\bibitem{kwon2023reward}
M.~Kwon, S.~M. Xie, K.~Bullard, and D.~Sadigh, ``Reward design with language models,'' in \emph{The Eleventh International Conference on Learning Representations}, 2023.

\bibitem{Bai2022TrainingAH}
Y.~Bai, A.~Jones, and K.~Ndousse, ``Training a helpful and harmless assistant with reinforcement learning from human feedback,'' \emph{ArXiv}, vol. abs/2204.05862, 2022.

\bibitem{ruan2024llm}
K.~Ruan, X.~Wang, and X.~Di, ``From twitter to reasoner: Understand mobility travel modes and sentiment using large language models,'' in \emph{2024 IEEE International Conference on Intelligent Transportation Systems (ITSC)}, 2024.

\bibitem{mao2023alympics}
S.~Mao, Y.~Cai, Y.~Xia, W.~Wu, X.~Wang, F.~Wang, T.~Ge, and F.~Wei, ``Alympics: Language agents meet game theory,'' 2023.

\bibitem{chahine2024large}
\BIBentryALTinterwordspacing
M.~Chahine, T.-H. Wang, H.~Zhang, W.~Xiao, D.~Rus, and C.~Gan, ``Large language models can design game-theoretic objectives for multi-agent planning,'' 2024. [Online]. Available: \url{https://openreview.net/forum?id=DnkCvB8iXR}
\BIBentrySTDinterwordspacing

\bibitem{horton2023llm}
J.~Horton, ``Large language models as simulated economic agents: What can we learn from homo silicus?'' \emph{SSRN Electronic Journal}, 01 2023.

\bibitem{akata2023playing}
E.~Akata, L.~Schulz, J.~Coda-Forno, S.~J. Oh, M.~Bethge, and E.~Schulz, ``Playing repeated games with large language models,'' 2023.

\bibitem{brand2023market}
J.~Brand, A.~Israeli, and D.~Ngwe, ``Using gpt for market research,'' \emph{SSRN Electronic Journal}, 2023.

\bibitem{yiting2023ration}
Y.~Chen, T.~X. Liu, Y.~Shan, and S.~Zhong, ``The emergence of economic rationality of gpt,'' \emph{Proceedings of the National Academy of Sciences}, vol. 120, no.~51, p. e2316205120, 2023.

\bibitem{dillion2023llm}
D.~Dillion, N.~Tandon, Y.~Gu, and K.~Gray, ``Can ai language models replace human participants?'' \emph{Trends in Cognitive Sciences}, vol.~27, no.~7, pp. 597--600, 2023.

\bibitem{aher2023llm}
G.~V. Aher, R.~I. Arriaga, and A.~T. Kalai, ``Using large language models to simulate multiple humans and replicate human subject studies,'' in \emph{Proceedings of the 40th International Conference on Machine Learning}, ser. Proceedings of Machine Learning Research, vol. 202.\hskip 1em plus 0.5em minus 0.4em\relax PMLR, 23--29 Jul 2023, pp. 337--371.

\bibitem{argyle2023lan}
L.~Argyle, E.~Busby, N.~Fulda, J.~Gubler, C.~Rytting, and D.~Wingate, ``Out of one, many: Using language models to simulate human samples,'' \emph{Political Analysis}, vol.~31, pp. 1--15, 02 2023.

\bibitem{sumers2024cognitive}
T.~R. Sumers, S.~Yao, K.~Narasimhan, and T.~L. Griffiths, ``Cognitive architectures for language agents,'' 2024.

\bibitem{xu2023exploring}
Y.~Xu, S.~Wang, P.~Li, F.~Luo, X.~Wang, W.~Liu, and Y.~Liu, ``Exploring large language models for communication games: An empirical study on werewolf,'' 2023.

\bibitem{Ouyang2022TrainingLM}
L.~Ouyang, J.~Wu, and X.~Jiang, ``Training language models to follow instructions with human feedback,'' \emph{NeurIPS}, 2022.

\bibitem{fan2023large}
C.~Fan, J.~Chen, Y.~Jin, and H.~He, ``Can large language models serve as rational players in game theory? a systematic analysis,'' 2023.

\bibitem{yin2023deep}
\BIBentryALTinterwordspacing
K.~Yin and A.~Mostafavi, ``Deep learning-driven community resilience rating based on intertwined socio-technical systems features,'' 2023. [Online]. Available: \url{https://arxiv.org/abs/2311.01661}
\BIBentrySTDinterwordspacing

\bibitem{Peng2024causal}
C.~Peng, D.~Zhang, and U.~Mitra, ``Graph identification and upper confidence evaluation for causal bandits with linear models,'' in \emph{ICASSP 2024 - 2024 IEEE International Conference on Acoustics, Speech and Signal Processing (ICASSP)}, 2024, pp. 7165--7169.

\bibitem{delgado2002conventions}
J.~Delgado, ``Emergence of social conventions in complex networks,'' \emph{Artificial Intelligence}, vol. 141, pp. 171--185, 10 2002.

\bibitem{sen2007social}
S.~Sen and S.~Airiau, ``Emergence of norms through social learning,'' in \emph{Proceedings of the 20th International Joint Conference on Artifical Intelligence}, ser. IJCAI'07.\hskip 1em plus 0.5em minus 0.4em\relax San Francisco, CA, USA: Morgan Kaufmann Publishers Inc, 2007, p. 1507–1512.

\bibitem{villatoro2011norm}
D.~Villatoro, J.~Sabater-Mir, and S.~Sen, ``Social instruments for robust convention emergence,'' in \emph{IJCAI International Joint Conference on Artificial Intelligence}, 01 2011, pp. 420--425.

\bibitem{yu2013collective}
C.~Yu, M.~Zhang, F.~Ren, and X.~Luo, ``Emergence of social norms through collective learning in networked agent societies,'' in \emph{Proceedings of the 2013 International Conference on Autonomous Agents and Multi-Agent Systems}, ser. AAMAS '13.\hskip 1em plus 0.5em minus 0.4em\relax Richland, SC: International Foundation for Autonomous Agents and Multiagent Systems, 2013, p. 475–482.

\bibitem{frank2013convention}
H.~Franks, N.~Griffiths, and A.~Jhumka, ``Manipulating convention emergence using influencer agents,'' \emph{Autonomous Agents and Multi-Agent Systems}, vol.~26, 05 2013.

\bibitem{lerer2019conventions}
\BIBentryALTinterwordspacing
A.~Lerer and A.~Peysakhovich, ``Learning existing social conventions via observationally augmented self-play,'' in \emph{Proceedings of the 2019 AAAI/ACM Conference on AI, Ethics, and Society}, ser. AIES '19.\hskip 1em plus 0.5em minus 0.4em\relax New York, NY, USA: Association for Computing Machinery, 2019, p. 107–114. [Online]. Available: \url{https://doi.org/10.1145/3306618.3314268}
\BIBentrySTDinterwordspacing

\bibitem{koster2020convention}
R.~Köster, K.~McKee, R.~Everett, L.~Weidinger, W.~Isaac, E.~Hughes, E.~Duenez-Guzman, T.~Graepel, M.~Botvinick, and J.~Leibo, \emph{Model-free conventions in multi-agent reinforcement learning with heterogeneous preferences}, 10 2020.

\bibitem{chen2022sl}
X.~Chen, Z.~Li, and X.~Di, ``Social learning in markov games: Empowering autonomous driving,'' in \emph{2022 IEEE Intelligent Vehicles Symposium (IV)}, 2022, pp. 478--483.

\bibitem{openai2024gpt4}
OpenAI, J.~Achiam, and et~al., ``{GPT-4 Technical Report},'' 2024.

\bibitem{michael1994markov}
M.~L. Littma, ``Markov games as a framework for multi-agent reinforcement learning,'' \emph{ICML}, pp. 157--–163, 1994.

\bibitem{vaswani2023attention}
A.~Vaswani, N.~Shazeer, N.~Parmar, J.~Uszkoreit, L.~Jones, A.~N. Gomez, L.~Kaiser, and I.~Polosukhin, ``Attention is all you need,'' 2023.

\bibitem{hadi2023large}
M.~U. Hadi, R.~Qureshi, A.~Shah, M.~Irfan, A.~Zafar, M.~B. Shaikh, N.~Akhtar, J.~Wu, S.~Mirjalili \emph{et~al.}, ``Large language models: a comprehensive survey of its applications, challenges, limitations, and future prospects,'' \emph{Authorea Preprints}, 2023.

\bibitem{mohapatra2023exploring}
H.~Mohapatra and S.~R. Mishra, ``Exploring ai tool's versatile responses: An in-depth analysis across different industries and its performance evaluation,'' 2023.

\bibitem{nyborg2016social}
K.~Nyborg, J.~M. Anderies, A.~Dannenberg, T.~Lindahl, C.~Schill, M.~Schl{\"u}ter, W.~N. Adger, K.~J. Arrow, S.~Barrett, S.~Carpenter \emph{et~al.}, ``Social norms as solutions,'' \emph{Science}, vol. 354, no. 6308, pp. 42--43, 2016.

\bibitem{openaiapi}
\BIBentryALTinterwordspacing
OpenAI. (2024) {OpenAI API Reference}. [Online]. Available: \url{https://platform.openai.com/docs/api-reference/chat}
\BIBentrySTDinterwordspacing

\bibitem{openaitokenizer}
\BIBentryALTinterwordspacing
------. (2024) {OpenAI Tokenizer}. [Online]. Available: \url{https://platform.openai.com/tokenizer}
\BIBentrySTDinterwordspacing

\end{thebibliography}

\end{document}